# Augmenting PID Control with Deep Reinforcement Learning: A Hybrid Approach to the Industrial Benchmark

Zhengyang (Cissy) Gu*
*Data Science*
*Liveline Technologies*
Livonia, MI, USA
cissy.gu@liveline.tech
0009-0007-8564-9907

Joseph E. Hernandez*
*Engineering*
*Liveline Technologies*
Livonia, MI, USA
joseph.hernandez@liveline.tech
0000-0002-2327-4034

John Burtenshaw
*Engineering*
*Liveline Technologies*
Livonia, MI, USA
john.burtenshaw@liveline.tech

Sean Scott
*Solutions*
*Liveline Technologies*
Livonia, MI, USA
sean.scott@liveline.tech

Thomas Cook
*Engineering*
*Liveline Technologies*
Livonia, MI, USA
thomas.cook@liveline.tech

Chris Couch
*Chief Executive Officer*
*Liveline Technologies*
Livonia, MI, USA
chris.couch@liveline.tech

***Abstract*—As industrial processes grow in complexity, traditional Proportional-Integral-Derivative (PID) controllers are often insufficient for handling their non-linear, multi-input dynamics. We propose using advanced Deep Reinforcement Learning (DRL) to prove its advantages in these complex environments. To do this, we rely on the Industrial Benchmark (IB) [1]. The IB is a realistic simulation that tests DRL algorithms against the key challenges of industrial applications: high-dimensional state spaces, delayed effects, and conflicting multi-criterial objectives. This testbed highlights DRL's core trade-off: while its final policies can often be unstable, its unique strength is the ability to autonomously discover optimal, non-obvious policies in multi-dimensional spaces where simple controllers fail.**

**In this paper, we propose a novel hybrid PID-RL controller that leverages DRL's discovery capability while ensuring Reliability. After developing a multi-objective reward function to make DRL viable, we use a twin-delayed deep deterministic (TD3) agent as a discovery tool to find the optimal, non-obvious settings for the IB's 'Gain' and 'Shift' parameters. By feeding these discovered parameters to a simple, tuned PID controller, our hybrid model successfully combines all three characteristics: it achieves the optimal Performance and Efficiency of the best DRL agent with the Reliability of a classical controller. This work demonstrates a practical methodology for using DRL to augment, rather than replace, trusted industrial control systems.**

## I. INTRODUCTION

Deep Reinforcement Learning (DRL) has shown immense promise for solving complex, high-dimensional control problems [2]-[4]. However, in industrial settings, DRL agents are often viewed with skepticism due to their "black box" nature, potential for instability, and high sample complexity. Conversely, the Proportional-Integral-Derivative (PID) controller remains the workhorse of industrial automation [5]. It is robust, interpretable, and well-understood, but it is fundamentally a Single-Input, Single-Output (SISO) controller and struggles with complex, Multi-Input, Multi-Output (MIMO) problems [6]-[8].

This paper explores a "best of both worlds" approach: a **Hybrid PID-RL controller**. We hypothesize that DRL's greatest value in industrial process control is not as an end-to-end replacement, but as a powerful discovery tool to find optimal configurations for simpler, trusted controllers [9].

To test this, we use the Industrial Benchmark (IB) [1]. While foundational DRL benchmarks like CartPole [10] or Atari[2] are useful, they lack the high-dimensional, continuous state-action spaces of real-world industry. The IB, in contrast, was designed to encapsulate these challenges, providing a far better testbed for DRL solutions that promise generalization to continuous process control.

The IB is intentionally designed with the difficulty of real-world applications. Its state-space is not only continuous and high-dimensional (180 dimensions) but also only **partially observable**, and its dynamics include significant **delayed effects**. Furthermore, the IB features a 3-dimensional continuous action space (Velocity, Gain, Shift) where the agent must balance its primary goal of setpoint tracking against conflicting secondary goals of consumption and fatigue. This creates a non-trivial challenge for simple controllers: while a PID can manage the setpoint using Velocity, it cannot optimize the efficiency, which is governed by the complex, coupled dynamics of Gain and Shift.

This project was funded by Liveline Technologies. Liveline Technologies is dedicated to improving manufacturing performance by harnessing the power of Artificial Intelligence to automate complex processes and predict future problems.

This is the author's version of this work. It is posted here for personal use, not for redistribution. The definitive version was published in the 2026 7th International Conference on Artificial Intelligence, Robotics and Control (AIRC), DOI: 10.1109/AIRC69745.2026.11631341.

Given this 3-dimensional challenge, our goal is to apply DRL. However, before DRL can be applied, we must first address the benchmark's flawed, cost-only reward function, which incentivizes inaction. We develop a multi-objective reward (Section III) that balances performance with efficiency. Using this, we evaluate several DRL agents (SAC [11], TD3 [12], DDPG [13]) and show that our hybrid model is the superior solution.

Our contributions are:

1) We propose a **Hybrid PID-RL controller** that uses a DRL agent to discover optimal system parameters (Gain, Shift) for a classical PID controller.
2) We demonstrate that DRL (specifically TD3) can be used as a **discovery tool** to find a non-obvious, 3-dimensional "eco-mode" policy that outperforms a naive PID.
3) We provide a comprehensive benchmark showing that DRL-only agents (TD3, DDPG) suffer from **bi-modal stability**, making them unreliable for production, whereas the Hybrid PID-RL model is robust.
4) We introduce the **necessary multi-objective reward shaping** and scaling required to make DRL agents successfully solve the IB.

## II. Background and Related Work

### A. The Industrial Benchmark (IB)

The goal of the Industrial Benchmark (IB) [1] is to provide a framework for developing advanced control methods to solve complex, multi-objective control problems. The agent must drive a system output (called 'Velocity (State)') from a start value (e.g., 50) to a target setpoint (e.g., 70), while simultaneously minimizing two conflicting physical costs: **Consumption Cost** (energy use) and **Fatigue Cost** (mechanical wear). To accomplish this, the agent has a 3-dimensional continuous action space, where each action $a \in [-1,1]$ controls one or few of the system's steerings. These actions directly influence the three core, observable system states:

- **Velocity (Action)**: The primary "gas pedal." It is used to drive the main system output ('Velocity (State)') towards the setpoint.
- **Gain (Action)**: Controls the system's sensitivity or responsiveness ('Gain (State)'). A high gain ("Sport Mode") is responsive but costs more energy, directly impacting the **Consumption Cost**.
- **Shift (Action)**: Adjusts an internal calibration ('Shift (State)'). An improper shift causes system stress, directly impacting the **Fatigue Cost**.

The IB is a generalizable model for many continuous control problems. As a **physics-based simulation**, its dynamics capture key features of real-world systems: (1) **Coupled, Multi-Input, Multi-Output (MIMO) control**, where all actions influence all outcomes; (2) **Conflicting objectives** (performance vs. efficiency vs. safety); and (3) **Time-delayed consequences and non-linear dynamics**. The observation space is also high-dimensional (180 dimensions) and partially observable, adding to the realism.

This structure makes it a solid analog for many industrial applications. For example:

- **Chemical Plants [7], [14]**: The 'Velocity (State)' is the product output. The 'Velocity (Action)' is the reagent flow, the 'Gain (Action)' is the reactor temperature (higher temp = faster reaction but higher *Consumption*), and the 'Shift (Action)' is the catalyst mixture (wrong mix = *Fatigue*).
- **Power Grid Management [6]**: The 'Velocity (State)' is the power output. The 'Velocity (Action)' is the generation command, the 'Gain (Action)' is the turbine ramp-up speed (fast ramp = high fuel *Consumption*), and the 'Shift (Action)' is the grid frequency (improper balance = grid *Fatigue*).
- **Robotic Manufacturing [3]**: The 'Velocity (State)' is the production rate. The 'Velocity (Action)' is the tool speed, the 'Gain (Action)' is the motor torque (high torque = high *Consumption*), and the 'Shift (Action)' is the tool pressure (wrong pressure = tool *Fatigue*).

### B. Deep Reinforcement Learning (DRL) for Continuous Control

Deep Reinforcement Learning (DRL) algorithms are broadly categorized as on-policy or off-policy. On-policy algorithms, such as A2C or PPO, are often sample-inefficient; they learn from an experience once and then discard it. This requires a massive number of new, live interactions to learn, which is expensive. Furthermore, their trial-and-error exploration must happen on the live system. In an industrial setting, this is often prohibitive, as a bad exploratory action could be dangerous or catastrophic [10].

For this reason, we focus on **off-policy** algorithms. These methods use a replay buffer, allowing them to "study" past experiences many times, making them far more sample-efficient. We evaluate three prominent off-policy, actor-critic algorithms [15] designed for continuous control:

- **DDPG (Deep Deterministic Policy Gradient)**: A foundational actor-critic method that learns a *deterministic* policy (it outputs a single, specific action, e.g., '0.51') [13]. The "Actor" network decides the action, and the "Critic" network learns to evaluate its quality (Q-value).
- **TD3 (Twin Delayed DDPG)**: A state-of-the-art successor to DDPG that addresses its well-known instability and Q-value overestimation [12]. It uses two "Twin" critic networks and takes the minimum of their predictions to prevent over-optimism. It also "delays" the update of the actor, which leads to much smoother and more stable training.
- **SAC (Soft Actor-Critic)**: A state-of-the-art method that learns a *stochastic* (probabilistic) policy [11]. It is optimized to maximize a trade-off between reward and policy entropy (randomness). This built-in exploration is very powerful, but as our results will show, its desire to be “random” can lead to unwanted oscillations in setpoint-tracking tasks.

### C. Reward Shaping

A reward function is the agent’s only specification of the goal [16]. Reward shaping is the process of modifying this function to guide the agent toward an optimal policy [17]. This is especially crucial in tasks with sparse rewards or, as in this case, misaligned reward signals [18]. Our approach can be viewed as a Multi-Objective Reinforcement Learning (MORL) problem [14], [19], where we define the true objective as a weighted sum of competing reward streams.

### D. PID Control

The Proportional-Integral-Derivative (PID) controller is the workhorse of industrial automation [5]. It calculates an action $u(t)$ based on the error $e(t) = Setpoint - CurrentValue$.

$$u(t) = K_p e(t) + K_i \int e(t)dt + K_d \frac{de(t)}{dt} \quad (1)$$

It is exceptionally robust for Single-Input, Single-Output (SISO) problems but is difficult to tune [20] and, in its basic form, cannot manage the complex, coupled dynamics of a 3-dimensional MIMO problem like the IB.

## III. Prerequisite: Refining a Reward Function

Our goal was to test Deep Reinforcement Learning (DRL) and hybrid controllers. We aimed to determine if a DRL agent could discover the optimal 3-dimensional policy (Velocity, Gain, and Shift) required to master the Industrial Benchmark (IB), a task beyond the capability of a simple 1D Proportional-Integral-Derivative (PID) controller. However, before DRL can be applied, we must first address the IB’s flawed reward function. The standard reward $R = -(Cost_{cons} + Cost_{fatigue})$ simply encourages inaction [21]. Conversely, a naive reward based only on error, $R = -(Setpoint - Output)^2$, fails by encouraging “bang-bang” oscillations, as it contains no penalty for control effort [22].

### A. A Multi-Objective, Scaled Reward

A successful policy must balance all three goals: (1) Minimize Error, (2) Minimize Consumption, and (3) Minimize Fatigue. Naively summing these fails, as their scales are vastly different (e.g., $Cost_{error} \sim 4900\ vs. Cost_{cons} \sim 400$). We therefore designed a scaled and weighted reward function.

Our final, tuned reward function is defined by the following equations:

$$R_{total} = -\left(w_e \cdot Cost_{error} + w_c \cdot Cost_{cons} + w_f \cdot Cost_{fatigue}\right) \quad (2)$$

This formulation defines the reward as the negative of a total, weighted cost. Therefore, the reward is a negative number, and the agent’s objective is to maximize this value. **A higher value (e.g., -200) is a better reward than a lower value (e.g., -500).** The agent achieves this by *minimizing* the combined costs.

Where the component costs are:

$$Cost_{error} = \left(\frac{Setpoint - Output}{10.0}\right)^2$$

$$Cost_{cons} = RawConsumption$$

$$Cost_{fatigue} = RawFatigue$$

And the tuned weights are:

- $w_e = 1.0$ (Primary Objective)
- $w_c = 0.005$ (Scaled to be a minor penalty)
- $w_f = 0.005$ (Scaled to be a minor penalty)

This formulation is defined as follows:

- $R_{total}$: The reward is the negative of a total, weighted cost.
- $Cost_{error}$: This is the primary objective. We use the squared error to heavily penalize large deviations from the setpoint. The error is scaled (divided by 10.0) before squaring to make the cost curve steep and create a strong incentive to converge.
- $Cost_{cons}$ & $Cost_{fatigue}$ : These are the raw physics costs. Their weights ($w_c = 0.005, w_f = 0.005$) are set very low to scale them down.

This weighting forces the agent to prioritize the setpoint (driven by the high $Cost_{error}$), using the minor physics costs as “tie-breakers” to find the most efficient and stable way to stay there.

## IV. Experimental Setup

We ran a comparative study between three DRL agents and two PID controllers.

### A. DRL Agents (SAC, TD3, DDPG)

We used the *d3rlpy* library [23] to test Soft Actor-Critic (SAC), Twin Delayed DDPG (TD3), and Deep Deterministic Policy Gradient (DDPG).

- **Training**: We ran 50 iterations for each algorithm. Each iteration consisted of a **10,000-step random warmup phase** to ensure sufficient exploration and fill the replay buffer with a wide range of data. This was followed by **25,000 online training steps**, bringing the total online interaction for each agent to 35,000 steps per iteration [10].
- **Hyperparameters**: We used a standard learning rate of $3 \times 10^{-4}$ for both actor and critic learning.
- **Scalers**: Using *StandardObservationScaler* and *StandardRewardScaler* was absolutely essential [24]. The 180-dim observation vector and our new reward function have values on different scales; normalization is mandatory for stable training.

### B. PID Baseline Controller

We implemented a classical PI controller (D-gain $K_d = 0$).

- **Tuning**: We performed a grid search over 100 combinations of $K_p$ and $K_i$ to find the pair that yielded the highest total reward.
- **Control**: The PID *only* controls the Velocity (Action 1).

- **Baseline**: For a fair, "naive" baseline, the **Gain and Shift actions are fixed to 0.0** (their neutral setting).

*C. Hybrid PID-RL Controller*

This controller tests our central hypothesis.

- **Hybridization**: The **Gain and Shift actions are fixed** to the optimal constant values discovered by our best TD3 agent: Gain = −1.0 and Shift = +1.0.
- **Tuning**: With the system now operating in this DRL-discovered "eco-mode," we performed a **new, separate grid search** over 100 combinations of $K_p$ and $K_i$ to find the optimal PID tuning for this specific hybrid configuration.
- **Control**: The re-tuned PID controls Velocity.

*D. Evaluation Metrics*

To account for the stochastic nature of DRL training, which is highly sensitive to random weight initialization and exploration, we cannot rely on a single training run. We therefore performed **50 independent training iterations** for each DRL agent. For each iteration, a new agent was trained from scratch for **35,000 online steps**.

Immediately following training, each of these 50 agents was evaluated on a single 200-step episode, and we recorded the following metrics from this evaluation episode:

1) **Total Episode Return**: The cumulative sum of all rewards obtained during the evaluation episode. This metric measures the *overall quality* of the agent's policy, capturing its ability to balance the primary goal (reaching the setpoint) with the secondary objectives (minimizing consumption and fatigue).
2) **Mean Final-10 Reward**: The average reward of the *last 10 steps* of the episode. This is our primary metric for *final stability* and steady-state performance. A high, non-oscillating value here indicates the agent successfully converged on the setpoint and is holding it efficiently, which is a key requirement for process control.
3) **Full Trajectory Plot**: A qualitative, visual analysis of the episode. This plot is essential for understanding *how* the agent achieves its reward. It allows us to diagnose its behavior (e.g., oscillating vs. stable) and discover its strategy (e.g., finding the "eco-mode").

This multi-iteration approach allows us to analyze the statistical reliability and convergence consistency of each algorithm, rather than relying on a single "lucky" or "unlucky" training.

## V. RESULTS AND DISCUSSION

*A. Defining a "Good" Controller*

Based on our multi-objective reward function (2), a "good" controller for this problem must demonstrate three key characteristics:

1) **Performance (Accuracy)**: It must successfully drive the system output to the 70 setpoint and hold it there.
2) **Efficiency (Optimality)**: It must discover and utilize the optimal, non-obvious 3D policy (i.e., the correct Gain and Shift) to minimize the physical costs of consumption and fatigue.
3) **Reliability (Stability)**: It must converge to this optimal policy consistently, without severe oscillations or critical sensitivity to random initialization.

We evaluated each controller type against these three criteria.

*B. Deep Reinforcement Learning (DRL) Algorithm Comparison: Policy Discovery*

We first compared the qualitative performance of representative iterations from Soft Actor-Critic (SAC), Deep Deterministic Policy Gradient (DDPG), and Twin Delayed DDPG (TD3), shown in Fig. 1.

- **SAC (Stochastic Policy)**: As shown in Fig. 1(a), the SAC agent's policy is less stable. While it eventually achieves **Performance** by converging to the setpoint, its path is inefficient and moderately erratic. This behavior is a known trade-off of SAC's core design, which optimizes for both maximum reward and maximum entropy (randomness), preventing the policy from fully settling. It fails on **Efficiency** and **Reliability**.
- **DDPG (Deterministic Policy)**: The DDPG agent (Fig. 1(b)) fails on **Performance**. It learns an aggressive "bang-bang" control policy, causing the system to overshoot and oscillate, never achieving a stable state. This is a well-documented failure mode of DDPG [13], where the critic network often overestimates the value of extreme actions, teaching the actor to slam the controls rather than finding a stable, intermediate solution [12].
- **TD3 (Stable Deterministic Policy)**: TD3 (Fig. 1(c)) was the only DRL agent capable of achieving both **Performance** and **Efficiency**, but only in its best iterations. This success is by design; its use of "twin" critics and delayed policy updates [12] prevents the severe critic overestimation that causes DDPG's instability. Its policy converges smoothly to the setpoint and, critically, it *discovers the optimal, 3-dimensional control strategy*. It learns to:
  1) Set the **Gain Action** to **-1.0** (the "Eco Mode").
  2) Set the **Shift Action** to **+1.0** (the optimal calibration point).
  3) Use the **Velocity Action** to settle into micro-oscillations around **0.0**.

This demonstrates TD3's *capability* to solve the full multi-objective problem: it can autonomously find non-obvious settings to minimize all costs.

*C. Multi-Iteration Stability Analysis: The Reliability Problem*

While Section 5.2 showed TD3's *capability* for Performance and Efficiency, our 50-iteration analysis revealed a critical flaw in its **Reliability**.

Analyzing the summary plots (Fig. 2) confirms this.

- **SAC** (Fig. 2(a)) shows high variance. The total returns are inconsistent, confirming it rarely finds a stable policy.

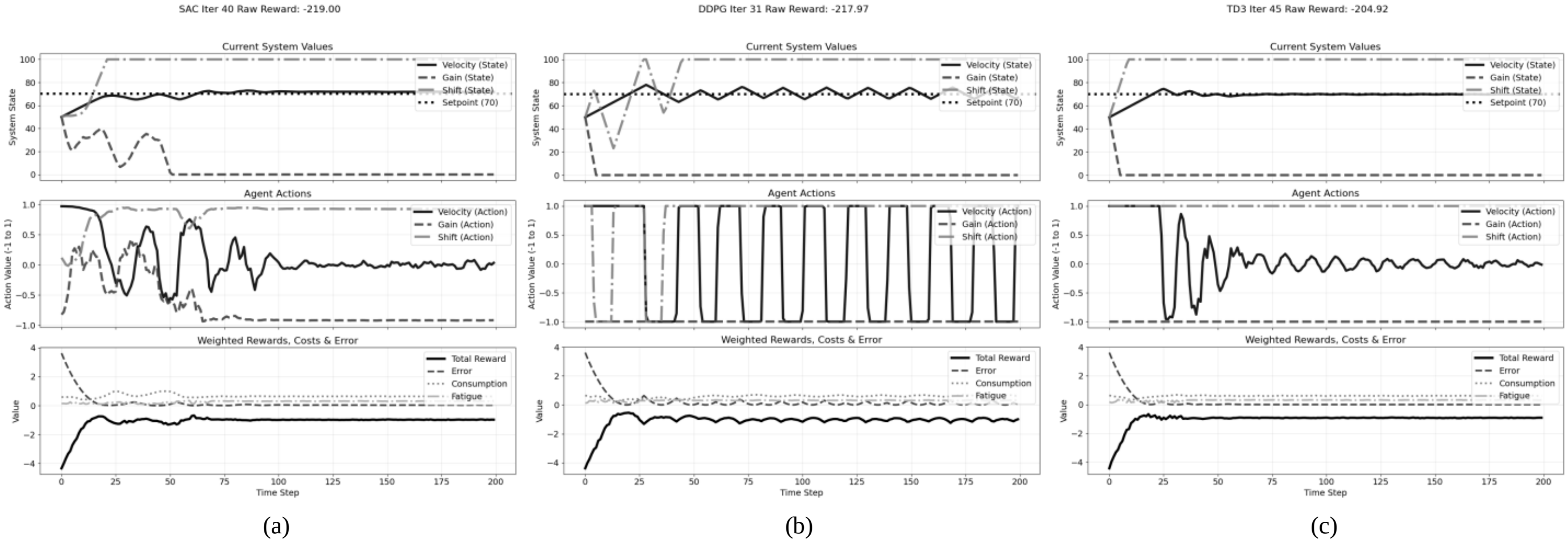

(a) (b) (c)

Fig. 1: Single-run evaluation of the three DRL agents. (a) SAC: Exhibits significant oscillation before eventually converging. (b) DDPG: Fails to converge, adopting an aggressive "bang-bang" control policy. (c) TD3: A successful iteration converges smoothly and discovers the optimal 3D policy: 'Gain =-1.0' and 'Shift = +1.0'.

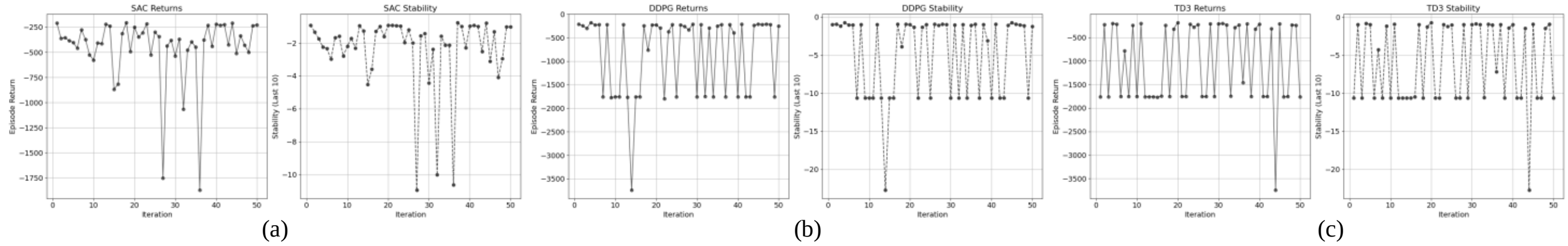

(a) (b) (c)

Fig. 2: Multi-iteration stability analysis for SAC, DDPG, and TD3 over 50 runs. (a) SAC shows high variance. (b, c) DDPG and TD3 show clear bimodal convergence (either succeeding or failing), highlighting their sensitivity to random initialization.

- **TD3** and **DDPG** (Fig. 2(b) and (c)) both exhibit a clear *bimodal* distribution. They either "solve" the problem (achieving a high, stable reward near 0) or get stuck in a bad local optimum (a very low reward).

This sensitivity to random initialization is the key problem. Fig. 3(c) shows a failed TD3 run (Iteration 50). Instead of finding the optimal policy, it converged to an incorrect local optimum, driving the output to 100. Its deterministic policy prevented it from exploring its way out.

This demonstrates that while DRL is a powerful discovery tool, its policies are brittle and lack the **Reliability** needed for a production environment.

### D. *Main Result: The Hybrid PID-RL as a Robust Solution*

This leads to our main experiment, which compares the DRL agents to our two PID controllers, shown in Fig. 3(a) and (b).

1) **Naive PID** (Fig. 3(a)): This controller achieves **Performance** (it reaches the setpoint) and is perfectly **Reliable**. However, it fails on **Efficiency**. Its "Standard Mode" (Gain=0.0) operation incurs a high Consumption cost, limiting its total reward to -291.56.
2) **Hybrid PID-RL** (Fig. 3(b)): This controller was re-tuned using the optimal parameters discovered by TD3 (Gain=-1.0, Shift=+1.0).

The results were conclusive: The **Hybrid PID-RL** model achieves a near-optimal reward of -203.46. This is statistically identical to the reward of the *best possible* autonomous TD3 agent (-204.92).

This is our key finding. The Hybrid model provides the **best of all three characteristics**:

- It achieves the **optimal Performance** and **Efficiency** of the best DRL agent.
- It provides the **Robustness** and **Reliability** of a simple PID controller, completely eliminating the bimodal failure risk seen in Fig. 3(c).

This confirms our hypothesis that DRL is most powerfully applied not as an end-to-end controller, but as an offline discovery tool to find optimal configurations for simpler, trusted systems.

## VI. Conclusion

The standard Industrial Benchmark (IB) is a powerful simulation, but its default reward function is flawed. It only incentivizes cost-savings, failing to encourage the agent to reach its actual goal.

We overcame this by developing a robust, multi-objective, and scaled reward function that successfully balances Performance (error) with Efficiency (consumption and fatigue).

With this new reward, we found that the deterministic Twin Delayed DDPG (TD3) agent, in its successful iterations, demonstrated high **Performance** by achieving stable setpoint tracking. Its key finding was the discovery of a complex, 3-action strategy (setting Gain to "eco-mode" and Shift to an optimal calibration point) that achieved superior **Efficiency**—a strategy a simple Proportional-Integral-Derivative (PID) controller could not discover.

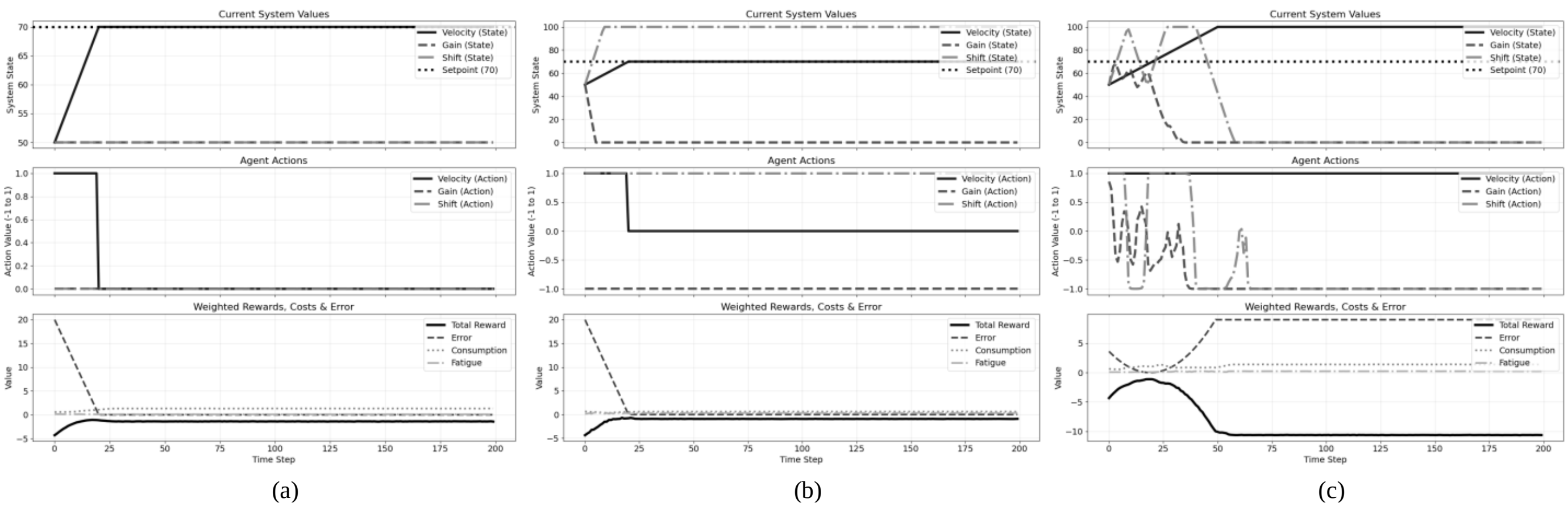

(a) (b) (c)

Fig. 3: Comparison of baseline, hybrid, and failed DRL runs. (a) Naive PID: Gain=0.0 (Reward:-291.56). (b) Hybrid PID RL: Uses DRL-discovered parameters (Gain=-1.0, Shift=+1.0) (Reward:-203.46). (c) Failed TD3: (Iter 50) Gets stuck in a local optimum, converging to 100.

However, our multi-iteration analysis showed that Deep Reinforcement Learning (DRL)-only approaches are brittle and lack **Reliability**. They suffer from bimodal stability, being highly sensitive to random initialization and prone to spectacular failures.

Our most significant contribution is the **Hybrid PID-RL controller**. This model provides the optimal solution by satisfying all three criteria: it achieves the **Performance** and **Efficiency** of the best-case DRL agent, but with the deterministic **Reliability** of a classical controller. By using the DRL agent as an **offline discovery tool** to find the optimal Gain and Shift parameters, and then "baking" those insights into a re-tuned PID, we achieved a robust, high-performing system. This suggests a practical path for AI in industry: use DRL to discover, then use PID to execute.